\documentclass[10pt,twocolumn,letterpaper]{article}

\usepackage[pagenumbers]{cvpr} 

\definecolor{cvprblue}{rgb}{0.21,0.49,0.74}
\usepackage[pagebackref,breaklinks,colorlinks,allcolors=cvprblue]{hyperref}

\usepackage[capitalize]{cleveref}
\crefname{section}{Sec.}{Secs.}
\Crefname{section}{Section}{Sections}
\Crefname{table}{Table}{Tables}
\crefname{table}{Tab.}{Tabs.}
\usepackage{bm}
\usepackage{capt-of}
\usepackage{cuted}
\usepackage{multirow}
\usepackage{relsize}
\usepackage{graphicx}
\usepackage{multirow}
\usepackage[ruled,vlined]{algorithm2e}
\usepackage{bbding}
\usepackage{wrapfig,lipsum,booktabs}
\usepackage{xcolor}
\usepackage{colortbl, booktabs}

\usepackage[normalem]{ulem}

\def\paperID{2736} 
\def\confName{CVPR}
\def\confYear{2025}

\title{Everything to the Synthetic: Diffusion-driven Test-time Adaptation via  Synthetic-Domain Alignment}
\author{%
Jiayi Guo$^{1,2}$\ \ \ Junhao Zhao$^1$\ \ \ Chaoqun Du$^1$\ \ \ Yulin Wang$^1$\ \ \ Chunjiang Ge$^1$\ \ \  Zanlin Ni$^1$\\\ Shiji Song$^1$ \ \ \ Humphrey Shi$^{2 *}$\ \ \ Gao Huang$^1$\thanks{Corresponding authors.} \\\
{\small$^1$Tsinghua University \ \ \
$^2$SHI Labs @ Georgia Tech} \\ 
\vspace{-30mm}
}

\begin{document}
\maketitle
\begin{strip}
    \centering
  \includegraphics[width=\textwidth]{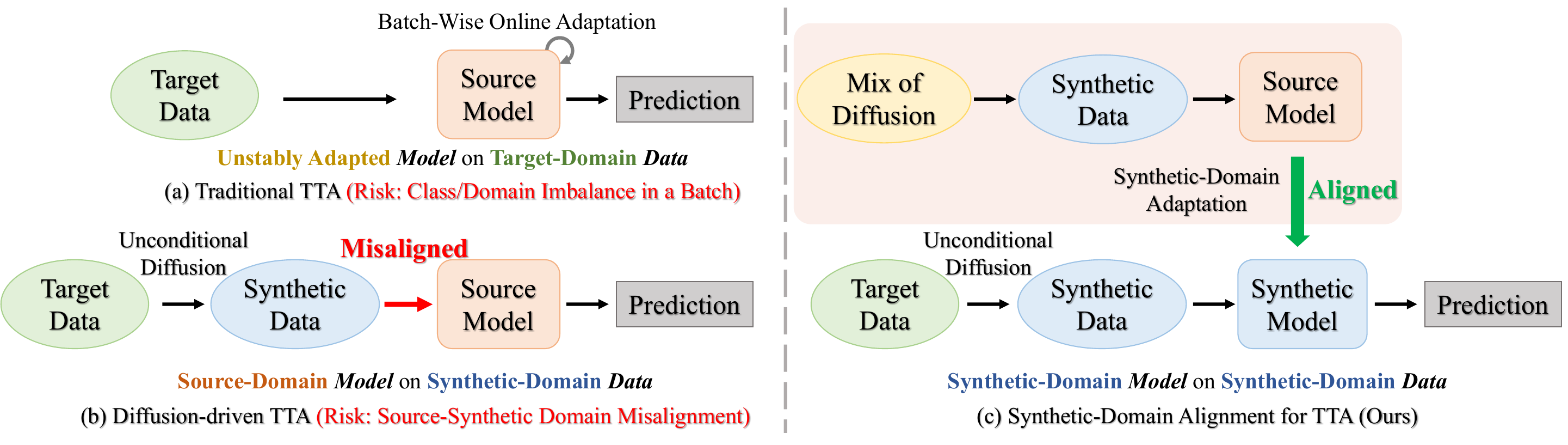}
    \vspace{-5mm}   
  \captionof{figure}{\textbf{Comparison of different test-time adaptation (TTA) frameworks.} (a) Traditional TTA methods continuously adapt source model weights to fit target data batches. However, their performance is sensitive to the amount and order of target data streams, \eg, adapting the model with batches containing data from only a single category can lead to overfitting. (b) Diffusion-driven TTA methods project the target data back to the synthetic domain of diffusion models, which still remains domain misalignment with the source domain. (c) We propose the Synthetic-domain Alignment (SDA) framework for TTA, which simultaneously aligns the domains of the source model and target data with the same synthetic domain for superior performance.}
  \label{fig:motivation}
    \vspace{-2mm}   

\end{strip}
\begin{abstract}
\vspace{-6mm}

Test-time adaptation (TTA) aims to improve the performance of source-domain pre-trained models on previously unseen, shifted target domains. Traditional TTA methods primarily adapt model weights based on target data streams, making model performance sensitive to the amount and order of target data. The recently proposed diffusion-driven TTA methods mitigate this by adapting model inputs instead of weights, where an unconditional diffusion model, trained on the source domain, transforms target-domain data into a synthetic domain that is expected to approximate the source domain. However, in this paper, we reveal that although the synthetic data in diffusion-driven TTA seems indistinguishable from the source data, it is unaligned with, or even markedly different from the latter for deep networks. To address this issue, we propose a \textbf{S}ynthetic-\textbf{D}omain \textbf{A}lignment (SDA) framework. Our key insight is to fine-tune the source model with synthetic data to ensure better alignment. Specifically, we first employ a conditional diffusion model to generate labeled samples, creating a synthetic dataset. Subsequently, we use the aforementioned unconditional diffusion model to add noise to and denoise each sample before fine-tuning. This Mix of Diffusion (MoD) process mitigates the potential domain misalignment between the conditional and unconditional models. Extensive experiments across classifiers, segmenters, and multimodal large language models (MLLMs, \eg, LLaVA) demonstrate that SDA achieves superior domain alignment and consistently outperforms existing diffusion-driven TTA methods. Our code is available at {{\href{https://github.com/SHI-Labs/Diffusion-Driven-Test-Time-Adaptation-via-Synthetic-Domain-Alignment}{https://github.com/SHI-Labs/Diffusion-Driven-Test-Time-Adaptation-via-Synthetic-Domain-Alignment}}}.

\vspace{-2mm}
\end{abstract}

\vspace{-2mm}
\section{Introduction}
\label{sec:intro}

Test-Time Adaptation (TTA)~\cite{bn1,tent,cotta,memo,ttt,maettt,diffpure,dda,gda} is an emerging research field that tackles domain misalignment when source models are evaluated on shifted target data. Unlike traditional domain adaptation (DA)~\cite{ganin2015unsupervised,saito2018maximum,long2018conditional} and source-free adaptation (SFA)~\cite{liang2020we,kundu2020universal,li2020model}, TTA addresses more practical scenarios where neither source data nor complete target data are accessible. Instead, the adaptation relies solely on streaming batches of target data.

\begin{table}[t]
\centering
\small
\resizebox{\linewidth}{!}{\begin{tabular}{ccc}
\toprule
  & Model Adaptation Direction & Data Adaptation Direction \\
  \midrule
  \multirow{2}{*}{Traditional TTA} & Expected: Source$\rightarrow$Target & \multirow{2}{*}{N/A}\\
  & \color{red}{Risk: Imbalanced Target Streams} & \\
  \midrule
  \multirow{2}{*}{Diffusion-driven TTA} & \multirow{2}{*}{N/A} & Expected: Target$\rightarrow$Source\\
  &&\color{red}{Actual: Target$\rightarrow$Synthetic}\\
  \midrule
  SDA (Ours) &Source$\rightarrow$Synthetic&Target$\rightarrow$Synthetic \\
\bottomrule 
\end{tabular}}\vspace{-1mm}
\caption{\textbf{Adaptation directions of different TTA methods.}}\label{tab:direct}\vspace{-6mm}
\end{table}

Traditional TTA methods (\cref{fig:motivation}a)~\cite{bn1,bn2,tent,cotta,memo,ttt,maettt,difftta} typically employ a \textit{source-to-target} model adaptation framework. These approaches continuously update the source model weights by processing target data batches. Without annotations, the adaptation process relies either on batch-wise updates of model statistics~\cite{bn1,bn2,tent,cotta}, or unsupervised or self-supervised auxiliary tasks~\cite{ttt,maettt,difftta}. However, small or imbalanced batches may poorly represent the target domain, making these approaches sensitive to the amount and order of the data stream~\cite{tent,dda,unitta}. For instance, adapting the model with batches containing data from only a single category can lead to overfitting.

Recently, the impressive generation capabilities of diffusion models \cite{ddpm,sd,dit} have sparked the development of diffusion-driven TTA methods (\cref{fig:motivation}b)~\cite{diffpure,dda,gda}, leveraging a \textit{target-to-source} framework. These approaches employ an unconditional diffusion model, pretrained on the source domain, aiming to project each target sample to the source domain independently. This enables the source model to make predictions without modifying its weights.
As a preliminary work, DiffPure \cite{diffpure} addresses adversarial perturbations by first applying a forward diffusion process, introducing a small amount of noise to the target data, followed by a reverse diffusion process to restore a clean image to approach the source domain. Building on this concept, more recent studies~\cite{dda,gda} tackle challenging domain shifts—such as severe data corruption—by incorporating additional structural guidance from the target data, helping preserve semantics and improve performance.

In this paper, we uncover that while diffusion-driven TTA methods aim to project target data back to the source domain, the projected target data remains confined within the synthetic domain of the unconditional diffusion model. 
As the synthetic-domain data are ultimately processed by the source-domain model, this domain misalignment limits the final performance. To address this issue, we propose Synthetic-Domain Alignment (\textbf{SDA}) (\cref{fig:motivation}c), a new category of framework for TTA tasks which \textit{simultaneously aligns the domains of the source model and target data with the same synthetic domain of a diffusion model} (\cref{tab:direct}). 

\begin{figure}[t]
  \centering
  \includegraphics[width=0.8\linewidth]{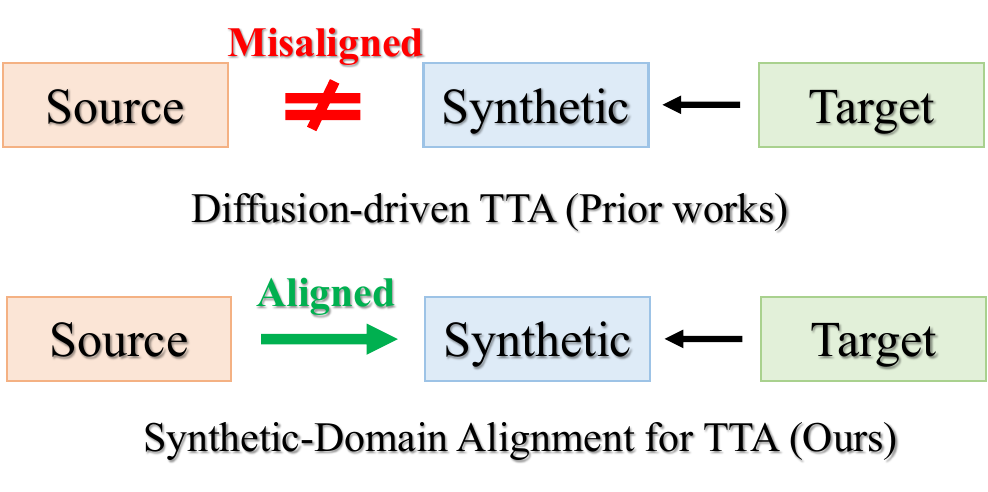}\vspace{-2mm}
  \caption{\textbf{Enhanced domain alignment with our framework}. Prior diffusion-driven TTA methods struggle with the domain misalignment between the source model and synthetic data, which we resolve by aligning the source model to the synthetic domain.}
  \vspace{-4mm}
  \label{fig:mo2}
\end{figure}

SDA distinguishes itself from existing diffusion-driven TTA methods~\cite{diffpure,dda,gda} by introducing an additional \textit{source-to-synthetic} model adaptation phase before testing on adapted target data (\cref{fig:mo2}). Since the adapted target data aligns with the synthetic domain generated by an unconditional diffusion model, SDA aims to adapt the source model to this same synthetic domain.
Specifically, SDA employs a Mix-of-Diffusion (\textbf{MoD}) technique to generate synthetic data for model adaptation. Given that the source data is inaccessible after pretraining, MoD first uses a conditional diffusion model to generate samples conditioned on domain-agnostic labels, creating a labeled synthetic dataset. Subsequently, the aforementioned unconditional diffusion model is leveraged by MoD add noise to and denoise these samples, addressing potential domain misalignment between the conditional and unconditional models. With a sufficiently large synthetic dataset, the fine-tuned model becomes highly effective at discriminating within the synthetic domain. Thus, the SDA framework transforms the cross-domain TTA task into an in-domain prediction task by aligning both the source model and target data with the same synthetic domain. 

SDA is a general framework, not limited to specific fine-tuning techniques or diffusion-driven data adaptation methods. This flexibility allows future advancements in these areas to further broaden its applicability. Extensive experiments across classifiers, segmenters, and multimodal large language models (MLLM, e.g., LLaVA) demonstrate that SDA consistently outperforms existing methods. Moreover, the effectiveness of our approach is reinforced through visualization analysis and ablation studies.

\section{Related Work}
\label{sec:related}

\noindent\textbf{Test-time adaptation (TTA)} is an emerging research area that addresses domain shifts by adapting either models~\cite{bn1,bn2,tent,cotta,memo,ttt,maettt,difftta} or data~\cite{diffpure,dda,gda} during evaluation on streaming target data batches. Early model adaptation methods update batch normalization statistics to match the target distribution~\cite{bn1,bn2,tent,memo}, while others leverage self-supervised tasks like rotation prediction~\cite{ttt} or image restoration~\cite{maettt,difftta} to adjust model weights. However, these approaches rely heavily on continuous weight updates, making them sensitive to the amount, order, and diversity of target data.
In contrast, diffusion-driven TTA methods~\cite{diffpure,dda,gda} focus on data adaptation by projecting each target sample back into the source domain, achieving stable performance without online model updates. DiffPure~\cite{diffpure} purifies adversarial samples with diffusion models, while DDA~\cite{dda} and GDA~\cite{gda} use structural guidance to preserve image content under severe corruption. Building on this, our work investigates the issue of misalignment between domains of the source model and synthetic images in diffusion-driven TTA and proposes a new synthetic-domain alignment TTA framework.

\noindent\textbf{Synthetic data for discriminative tasks.}
Synthetic data, generated by models rather than collected from the real world, has shown significant potential in enhancing visual representations for various discriminative tasks~\cite{tian2023learning,stablerep,scaling}. It has been effectively applied in areas such as visual recognition~\cite{azizi2023synthetic,stablerep}, object detection~\cite{detection1,detection2}, semantic segmentation~\cite{segmentation1,segmentation2,sankaranarayanan2018learning}, image assessment~\cite{risa}, autonomous driving~\cite{auto1}, and robotics~\cite{robo1,robo2}. In this work, we explore the potential of leveraging synthetic data, generated by diffusion models, for domain alignment in TTA tasks.

\begin{figure}[t]
  \centering
  \includegraphics[width=\linewidth]{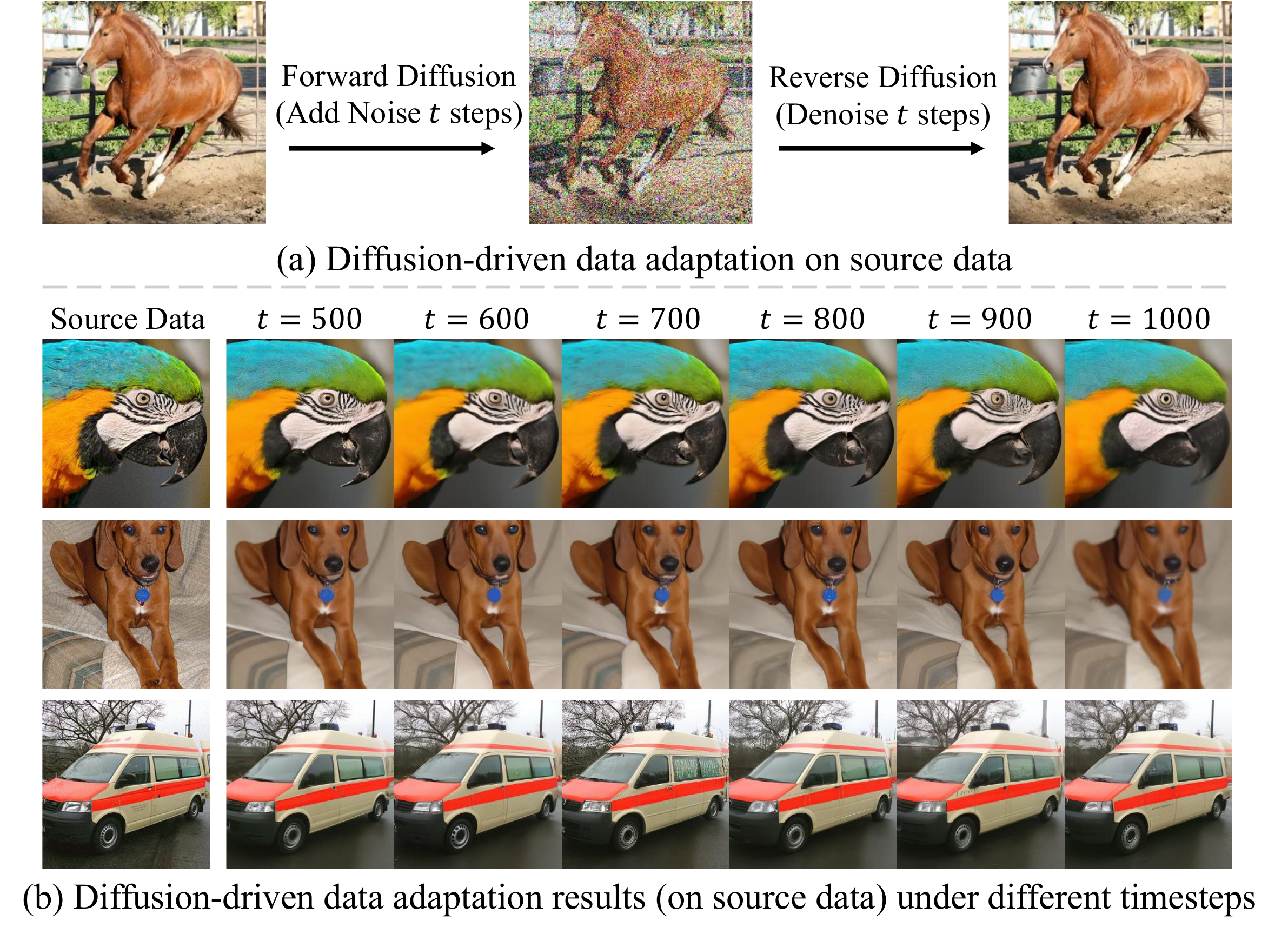}
  \vspace{-6mm}
  \caption{\textbf{(a) Illustration of diffusion-driven data adaptation on source data and (b) Adapted images across different timesteps.} The results are obtained using DDA~\cite{dda}, with no noticeable visual degradation observed in the adapted images.}
  \vspace{-4mm}
  \label{fig:pretest}
\end{figure}

\section{Methodology}
\label{sec:3-1}

In this section, we introduce the background of diffusion-driven TTA methods and identify their source-synthetic domain misalignment issue in~\cref{sec:bg} and~\cref{sec:mis}. To tackle this issue, we propose the synthetic-domain alignment (SDA) framework in~\cref{sec:sdaf} and introduce its key technique, the mix of diffusion (MoD) in~\cref{sec:mod}.

\begin{figure*}[t]
  \centering
  \includegraphics[width=0.95\linewidth]{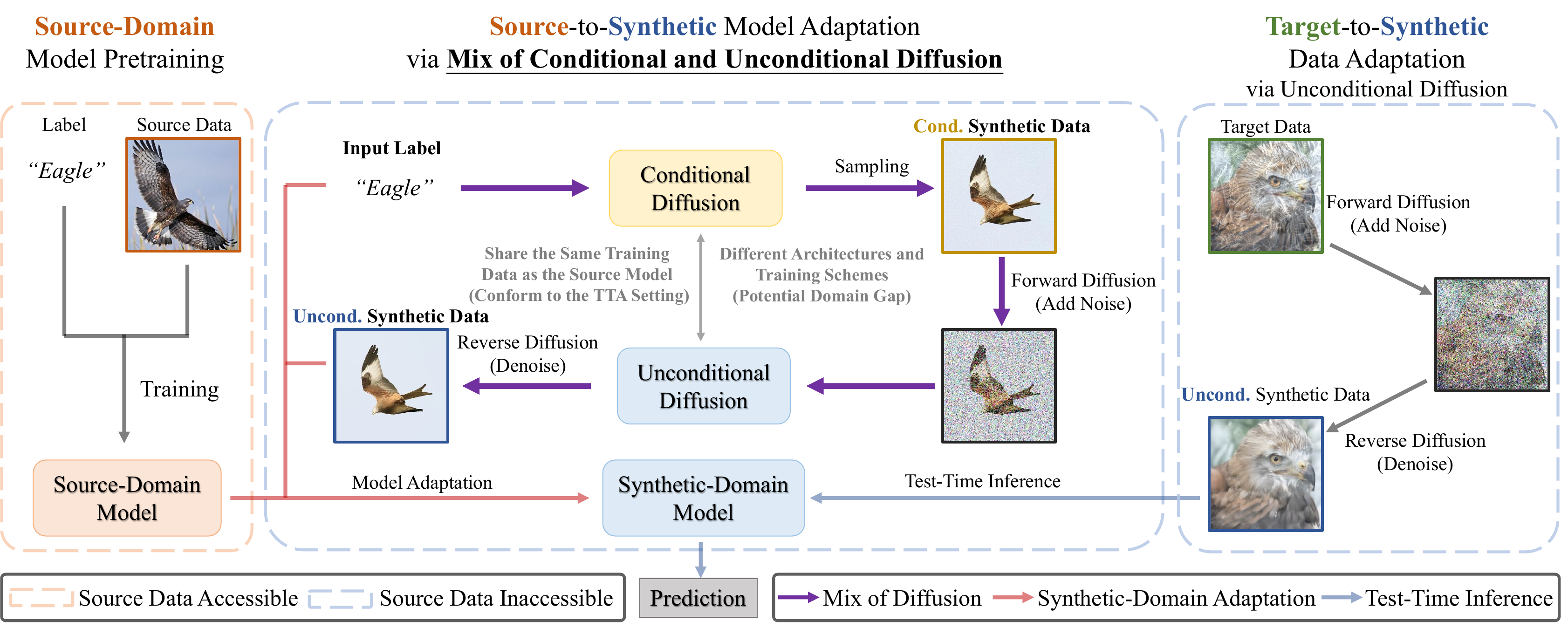}\vspace{-1mm}
  \caption{\textbf{Overview of the Synthetic-Domain Alignment (SDA) framework.} SDA is a novel TTA framework aligning both the domains of the source model and the target data with the synthetic domain. SDA involves three phases: (left): a \textit{source-domain model pretraining} phase, where the source model is trained on source data prior to TTA; (middle): a \textit{source-to-synthetic model adaptation} phase, where the source model is adapted to a synthetic-domain model using synthetic data generated via a Mix of Diffusion (MoD) technique; and (right): a \textit{target-to-synthetic data adaptation} phase, where target data is adapted into synthetic data using an unconditional diffusion model. Finally, the adapted synthetic data is fed into the synthetic-domain model for test-time inference.}
  \vspace{-3mm}
  \label{fig:method}
\end{figure*}

\subsection{Background}\label{sec:bg}

\textbf{Diffusion process.} Given a source data point $\bm{x}_0^{\rm src}\sim p_0^{\rm src}$, diffusion models $\epsilon_\theta$~\cite{ddpm} gradually transform $p_0^{\rm src}$ into a Gaussian noise distribution $N(\bm{0}, \bm{I})$ through a $T$-step forward diffusion process.  At each timestep ${t}\in \{1,2,\cdots,T\}$, the intermediate state $\bm{x}_t^{\rm src}\sim p_t^{\rm src}$ is computed as:
\begin{equation}
    \bm{x}_t^{\rm src} = \sqrt{1-\beta_t}\bm{x}_{t-1}^{\rm src} + \sqrt{\beta_t} \bm{\epsilon}_t,
\label{eq:xt}
\end{equation}
where $\bm{\epsilon}_t$ is random Gaussian noise and $\beta_t \in (0, 1)$ represents the diffusion rate at step $t$. By defining $\alpha_t = 1-\beta_t$, $\overline{\alpha_t} = \prod_{t=1}^{T}\alpha_t$ and $\bm{\epsilon} \sim N(\bm{0}, \bm{I})$, we obtain:
\begin{equation}
    \bm{x}_t^{\rm src} 
     = \sqrt{\overline{\alpha_t}}\bm{x}_{0}^{\rm src} + \sqrt{1-\overline{\alpha_t}} \bm{\epsilon}.
\label{eq:xt2}
\end{equation}

The reverse diffusion process recovers a clean $\widehat{\bm{x}_0^{\rm src}}$ by progressively removing noise from $\bm{x}_T^{\rm src}$:
\begin{equation}
    \widehat{\bm{x}_{t-1}^{\rm src}} = 
    \frac{1}{\sqrt{\alpha_t}}(\widehat{\bm{x}_{t}^{\rm src}}-\frac{1-\alpha_t}{\sqrt{1-\overline{\alpha_t}}}\epsilon_\theta(\widehat{\bm{x}_t^{\rm src}}, t)) + \sigma_t \bm{\epsilon},
\label{eq:xt-1}
\end{equation}
where $\sigma_t$ is the posterior noise variance~\cite{ddpm}. 
\vspace{1mm}

\begin{table}[t]
\centering
\small
\vspace{-1mm}
\setlength{\tabcolsep}{3pt}
\vspace{-2mm}
\resizebox{\linewidth}{!}{\begin{tabular}{l|cccc cc}
\toprule
Timestep $t$ 
& 500 & 600 & 700 & 800 & 900 & 1000\\
\midrule
Swin-B \textsubscript{(Source-Synthetic, {\color{red}{Misaligned}})} & 61.6 & 57.8 & 55.7 & 52.6 & 45.0 & 38.5 \\
\rowcolor{gray!20}Swin-B \textsubscript{(Synthetic-Synthetic, {\color{teal}{Aligned}})}   & 67.6  & 65.9  & 65.0  & 61.5  & 55.7  & 48.4\\ 
$\Delta$ & \color{teal}{+6.0} & \color{teal}{+8.1} & \color{teal}{+9.3} & \color{teal}{+8.9} & \color{teal}{+10.7} & \color{teal}{+9.9}\\
\midrule

ConvNeXt-B \textsubscript{(Source-Synthetic, {\color{red}{Misaligned}})} & 65.1 & 61.3 & 60.3 & 57.2 & 50.0 & 41.5\\
\rowcolor{gray!20}ConvNeXt-B \textsubscript{(Synthetic-Synthetic, {\color{teal}{Aligned}})}   & 70.6  & 68.2  & 67.4  & 64.9  & 58.5  & 50.7 \\
$\Delta$ & \color{teal}{+5.5} & \color{teal}{+6.9} & \color{teal}{+7.1} & \color{teal}{+7.7} & \color{teal}{+8.5} & \color{teal}{+9.2}\\

\bottomrule 
\end{tabular}}\vspace{-2mm}
\caption{\textbf{Source model accuracy across different timesteps of diffusion-driven data adaptation.} For suitable timesteps for TTA ($t\geq 500$)~\cite{dda}, model accuracy shows a monotonically decreasing trend with the growth of timestep, indicating the increase of the misalignment of source domain $p_0^{\rm src}$ and synthetic domain $p_{0, \text{u}}^{\rm syn}$. By aligning the source model to the synthetic domain, our methods (rows 3 \& 6) significantly help alleviate the performance degradation. Results are evaluated on the ImageNet~\cite{in} validation set.}\label{tab:pre}\vspace{-3mm}
\end{table}

\noindent\textbf{Diffusion-driven data adaptation.}
Denote by $p_0^{\rm trg}$ the target data distribution, from which each target data point $\bm{x}_t^{\rm trg}$ is sampled. Prior work~\cite{diffpure} demonstrates that:
\begin{equation}
    D_{\text{KL}}(p_{t+1}^{\rm src} || p_{t+1}^{\rm trg}) - D_{\text{KL}}(p_{t}^{\rm src} || p_{t}^{\rm trg}) \leq 0,
\label{eq:t*}
\end{equation}
where $D_{\text{KL}}$ is the KL divergence. Since $p_{T}^{\rm src} = p_{T}^{\rm trg} = N(\bm{0}, \bm{I})$,
for any arbitrarily small value $\delta$, there exists a minimum timestep $t^*$ such that $D_{\text{KL}}(p_{t^*}^{\rm src} || p_{t^*}^{\rm trg}) < \delta$. As discussed in~\cite{dda}, a diffusion process with $T=1000$ requires $t^*\geq 500$ to eliminate domain shifts. We empirically validate this choice of $t^*$ in the supplementary materials.

The initial diffusion-driven data adaptation in DiffPure~\cite{diffpure} is carried out as follows: Given the minor divergence between $p_{t^*}^{\rm src}$ and $p_{t^*}^{\rm trg}$, we recover each $\bm{x}_{t^*}^{\rm trg}\sim p_{t^*}^{\rm trg}$ to its corresponding $\widehat{\bm{x}_0^{\rm src}}\sim p_{0}^{\rm src}$ by executing the reverse diffusion process (Eq.~\ref{eq:xt-1}) $t^*$ times.

Since the reverse process is stochastic, subsequent works~\cite{dda,gda} further introduce additional structure guidance to ensure the content consistency between each $\bm{x}_{0}^{\rm trg}$ and its adapted $\widetilde{\bm{x}_0^{\rm src}}$. In this work, we adopt the same data adaptation process as DDA~\cite{dda}:
\begin{equation}
    \widetilde{\bm{x}_{t-1}^{\rm src}} = \widehat{\bm{x}_{t-1}^{\rm src}} - \bm{w} \nabla_{\widetilde{\bm{x}_{t}^{\rm src}}} \left\lVert \phi(\bm{x}_0^{\rm trg}) - \phi(\widetilde{\bm{x}_{0,t}^{\rm src}}) \right\rVert_2,
    \label{eq:dda}
\end{equation}
where $\widehat{\bm{x}_{t-1}^{\rm src}}$ is computed as Eq.~\ref{eq:xt-1}, $\bm{w}$ is the structure guidance scale, $\phi$ is a structure extractor~\cite{dda} and $\widetilde{\bm{x}_{0,t}^{\rm src}}$ is an estimate of $\widetilde{\bm{x}_{0}^{\rm src}}$ at timestep $t$~\cite{ddpm}. Diffusion-driven data adaptation is typically performed with unconditional diffusion models~\cite{diffusion-beats-gans} since the target data labels are unknown. 

\subsection{Source-Synthetic Domain Misalignment}\label{sec:mis}

Unlike a \textit{real} source data point \( \bm{x}_{0}^{\text{src}} \), the recovered version \( \widetilde{\bm{x}_{0}^{\text{src}}} \), derived from \( \bm{x}_{t^*}^{\text{trg}} \) is \textit{synthetic}. Specifically, \( \bm{x}_{0}^{\text{src}} \) follows the source domain \( p_0^{\text{src}} \), whereas \( \widetilde{\bm{x}_{0}^{\text{src}}} \) follows the synthetic domain \( p_{0, \text{u}}^{\text{syn}} \) of an unconditional diffusion model \( \epsilon_\theta^{\text{u}} \) with parameters $\theta$.
In this section, we empirically reveal the misalignment between the source domain 
\( p_0^{\text{src}} \) and the synthetic domain \( p_{0, \text{u}}^{\text{syn}} \),
and investigate how this misalignment impacts the performance of the source model. For simplicity and clarity, we will substitute \( \widetilde{\bm{x}_{0}^{\text{src}}} \) with \( \bm{x}_{0, \text{u}}^{\text{syn}} \) throughout the following discussion.  

Based on the above analysis, diffusion-driven TTA methods face two potential misalignments: (1) source-target domain misalignment arising from inherent data distribution shifts, which has been the primary focus of prior research, and (2) source-synthetic domain misalignment introduced by diffusion models, which we address as the main focus of this work, complementing existing efforts.

To precisely examine the impact of the source-synthetic domain misalignment and isolate it from the influence of the source-target domain misalignment, we evaluate the performance of the source model $f$ on synthetic data $\bm{x}_{0, {\rm u}}^{\rm syn}$ adapted by the diffusion model $\epsilon_\theta^{\rm u}$ from source data $\bm{x}_{0}^{\rm src}$ (\cref{fig:pretest}a).
Specifically, we test ImageNet pretrained models~\cite{in} on the ImageNet validation set adapted by the popular diffusion-driven data adaptation method, DDA~\cite{dda} across different timesteps $t$. As indicated in~\cref{tab:pre}, with an increase in $t$, the accuracy of the model exhibits a monotonically decreasing trend. For ideal $t^*\geq500$ for TTA tasks, performance degradation of more than 18.8\% for ConvNeXt~\cite{convnext} and 21.8\% for Swin~\cite{swin} is observed compared to their official results~\cite{2023mmpretrain} on source data which are 83.4\% and 83.9\%. This indicates a significant domain misalignment between $p_0^{\rm src}$ and $p_{0, \text{u}}^{\rm syn}$. By fine-tuning the source model on the synthetic data generated by our Mix of Diffusion process (introduced in~\cref{sec:mod}), the aligned models $f^\prime$ achieve performance improvements of 5.5\% for ConvNeXt and 6.0\% for Swin.

In~\cref{fig:pretest}b, we show that the diffusion synthetic data $\bm{x}_{0, {\rm u}}^{\rm syn}$ and source data $\bm{x}_{0}^{\rm src}$ exhibit no noticeable visual differences across different timesteps $t$. This further suggests that the performance degradation is not due to the quality of diffusion-generated images but rather the implicit misalignment between the source and synthetic domains.

\subsection{Synthetic-Domain Alignment Framework}\label{sec:sdaf}

Given that the diffusion-adapted data $\bm{x}_{0,{\rm u}}^{\rm syn}$ aligns more closely with the synthetic domain $p_{0, {\rm u}}^{\rm syn}$ instead of the source domain \( p_0^{\text{src}} \), we propose simultaneously adapting the source model $f$ to the same synthetic domain $p_{0, {\rm u}}^{\rm syn}$. By doing so, the alignment between the data and model within $p_{0, {\rm u}}^{\rm syn}$ can be effectively achieved. To this end, we introduce a novel TTA framework: \textbf{S}ynthetic-\textbf{D}omain \textbf{A}lignment (\textbf{SDA}).

In~\cref{fig:method}, we present the complete diagram of SDA, which consists of three key phases: (1) a \textit{source-domain model pretraining} phase, where the source model \( f \) is trained on source data \( \bm{x}_0^{\rm src} \) prior to TTA; (2) a \textit{source-to-synthetic model adaptation} phase, where \( f \) is fine-tuned to a synthetic-domain model \( f' \); and (3) a \textit{target-to-synthetic data adaptation} phase, where target data \( \bm{x}_0^{\rm trg} \) is adapted into synthetic data \( \bm{x}_{0,{\rm u}}^{\rm syn} \) using an unconditional diffusion model \( \epsilon_\theta^{\rm u} \) following Eq.~\ref{eq:dda}. Finally, the adapted synthetic data \( \bm{x}_{0,{\rm u}}^{\rm syn} \) is fed into the synthetic-domain model \( f' \) for test-time inference. 
Consistent with the standard TTA protocol~\cite{tent}, the source data is accessible only during the source model pretraining phase and remains inaccessible during both the model and data adaptation phases.

The rationale behind SDA is straightforward: by aligning the model and data domains to the same synthetic domain, the original \textit{cross-domain} TTA task is transformed into an easier \textit{in-domain} prediction task, thus addressing the core challenge of TTA and improving performance.

Since the pretraining phase follows standard supervised learning, and the data adaptation phase aligns with DDA~\cite{dda}, we omit detailed explanations of these phases. Instead, we focus on how we adapt the source model \( f \) to an ideal synthetic domain model \( f' \) using a novel technique called Mix-of-Diffusion (\textbf{MoD}), introduced next.

\subsection{Model Adaptation via Mix of Diffusion}\label{sec:mod}

As shown in~\cref{fig:method}, MoD consists of two main processes: a data generation process powered by a conditional diffusion model  $\epsilon_\eta^{\rm c}$ with parameters $\eta$, and a data alignment process powered by the same unconditional diffusion model $\epsilon_\theta^{\rm u}$ used in the target-to-synthetic data adaptation phase. It is worth noting that, both $\epsilon_\eta^{\rm c}$ and $\epsilon_\theta^{\rm u}$ are also pretrained on source data $\bm{x}_{0}^{\rm src}$ and have never been exposed to target data $\bm{x}_{0}^{\rm trg}$, in accordance with the TTA setting.

\vspace{1mm}
\noindent\textbf{Conditional diffusion data generation.}
The generation process leverages the conditional generation capability of $\epsilon_\eta^{\rm c}$ to synthesize conditional synthetic data $\bm{x}_{0,{\rm c}}^{\rm syn} \sim p_{0,{\rm c}}^{\rm syn}$. In the context of TTA, the source and target domains share the same domain-agnostic label set $\{y_i\}_{i=1}^{K}$. Utilizing $\epsilon_\eta^{\rm c}$, we uniformly generate samples for each class $y_i$ from Gaussian noise (${\bm{x}_{T,c}^{\rm syn}}$) through a $T$-step reverse diffusion process: 
\begin{equation}
    {\bm{x}_{t-1,c}^{\rm syn}} = 
    \frac{1}{\sqrt{\alpha_t}}({\bm{x}_{t,c}^{\rm syn}}-\frac{1-\alpha_t}{\sqrt{1-\overline{\alpha_t}}}\epsilon_\eta^{\rm c}({\bm{x}_{t,c}^{\rm syn}}, t, y_i)) + \sigma_t \bm{\epsilon}.
\label{eq:x0c}
\end{equation}

The generation capability of $\epsilon_\eta^{\rm c}$ allows for the construction of a labeled synthetic-domain dataset $\{\bm{x}_{0,{\rm c}}^{\rm syn}, y\}^{N}$ of arbitrary size $N$ without any manual data collection. By fine-tuning the source model $f$ on this synthetic dataset, an adapted model $f_{\rm c}^\prime$ on domain $p_{0,{\rm c}}^{\rm syn}$ can be obtained.

\vspace{1mm}
\noindent\textbf{Unconditional diffusion data alignment.} However, since the test-time adapted data $\bm{x}_{0,{\rm u}}^{\rm syn}$ is subject to domain $p_{0,{\rm u}}^{\rm syn}$, we argue that there is still potential misalignment between different synthetic domains $p_{0,{\rm c}}^{\rm syn}$ and $p_{0,{\rm u}}^{\rm syn}$. This is mainly because of the differences in architectures and training schemes between $\epsilon_\eta^{\rm c}$ and $\epsilon_\theta^{\rm u}$. We empirically validate this argument in~\cref{table:abc}, which indicates that the model $f_{\rm c}^\prime$ adapted to domain $p_{0,{\rm c}}^{\rm syn}$ performs worse than the model $f_{\rm u}^\prime$ adapted to domain $p_{0,{\rm u}}^{\rm syn}$ during testing. 

To obtain the $f_{\rm u}^\prime$ on domain $p_{0,{\rm u}}^{\rm syn}$, we mirror the target-to-synthetic data adaptation phase as a conditional synthetic data ($\bm{x}_{0,{\rm c}}^{\rm syn}$) to unconditional synthetic data ($\bm{x}_{0,{\rm u}}^{\rm syn}$) adaptation process. In specific, we use the same $t^*$ in~\cref{sec:bg} with Eq.~\ref{eq:xt2} to obtain $\bm{x}_{t^*,{\rm c}}^{\rm syn}$. According to the analysis in~\cref{sec:bg}, $\bm{x}_{t^*,{\rm c}}^{\rm syn}$ will be indistinguishable to its counterpart $\bm{x}_{t^*,{\rm u}}^{\rm syn}$ on domain $p_{t^*,{\rm u}}^{\rm syn}$. Therefore, we have:
\begin{equation}
    \bm{x}_{t^*,{\rm u}}^{\rm syn} \approx  \bm{x}_{t^*,{\rm c}}^{\rm syn}
     = \sqrt{\overline{\alpha_t}}\bm{x}_{0,c}^{\rm syn} + \sqrt{1-\overline{\alpha_t}} \bm{\epsilon}.
\end{equation}

Then, following Eq.~\ref{eq:xt-1} and Eq.~\ref{eq:dda}, the noisy $\bm{x}_{t^*,{\rm u}}^{\rm syn}$ can be gradually denoised to $\bm{x}_{0,{\rm u}}^{\rm syn}$. Finally, the expected synthetic-domain model $f_{\rm u}^\prime$ is obtained by fine-tuning the source-domain model $f_{\rm u}^\prime$ on dataset $\{\bm{x}_{0,{\rm u}}^{\rm syn}, y\}^{N}$.

\vspace{1mm}
\noindent\textbf{Ensembling.} As noted in previous diffusion-driven TTA methods~\cite{dda}, while diffusion models generally perform well for data adaptation, they may occasionally produce data points that are less recognizable than the original target data. To address this, prior approaches use an ensemble of model predictions on \( \bm{x}_0^{\rm trg} \) and \( \bm{x}_{0,{\rm u}}^{\rm syn} \) as the final output. Following this protocol, the final prediction in SDA is:

\begin{equation}
    \hat{y}=\arg \max_y (q(y|\bm{x}_0^{\rm trg}) + q^\prime(y|\bm{x}_{0,{\rm u}}^{\rm syn})),
\label{eq:ensemble}
\end{equation}
where $q(\cdot)$ and $q^\prime(\cdot)$ are output distributions of source model $f$ and synthetic-domain model $f_{\rm u}^\prime$, respectively.

\section{Experiments}
\label{sec:expr}
In this section, we first evaluate SDA on ImageNet classifiers with standard TTA benchmarks in~\cref{sec:expin}. Next, we assess SDA’s scalability across different dataset sizes, tasks, and model architectures in~\cref{sec:expscale}. In~\cref{sec:expan}, we demonstrate SDA’s advantages through Grad-CAM visualizations~\cite{gradcam} and data stream sensitivity tests. Finally, ablation studies in~\cref{expab} validate the design choices in our SDA framework.
\begin{table}[h]
\centering
\small
\setlength{\tabcolsep}{3pt}
\resizebox{\linewidth}{!}{\begin{tabular}{ccccc|c>{\columncolor{gray!20}}c}
\toprule
 Model& Source & MEMO & DiffPure & GDA & DDA   & SDA (Ours) \\
\midrule
ResNet-50     & 18.7        & 24.7 & 16.8     &  31.8 &   29.7 &\textbf{32.5 \color{teal}{(+2.8)}}        \\
Swin-T        & 33.5        & 29.5 & 24.8     & 42.2 &  40.0 & \textbf{42.5 \color{teal}{(+2.5)}}          \\
ConvNeXt-T      & 39.3        & 37.8 & 28.8     & 44.8 &  44.2 &  \textbf{47.0 \color{teal}{(+2.8)}}         \\
Swin-B         & 40.5        & 37.0 & 28.9     &  -    &   44.5 &   \textbf{47.4 \color{teal}{(+2.9)}}      \\
ConvNeXt-B    & 45.6        & 45.8 & 32.7     &  -    & 49.4 & \textbf{51.9 \color{teal}{(+2.5)}} \\   
\bottomrule 
\end{tabular}}
\vspace{-0mm}
\caption{\textbf{Comparison results on ImageNet-C~\cite{inc}}. We compare SDA with source models, MEMO~\cite{memo}, DiffPure~\cite{diffpure}, GDA~\cite{gda} and DDA~\cite{dda}. Results are the average accuracy across 15 adaptation domains at severity level 5. SDA shows consistent performance improvements compared to baselines.}\label{tab:expinc}\vspace{-2mm}
\end{table}

\begin{table*}[t]
\centering
\small
\vspace{-1mm}
\setlength{\tabcolsep}{3pt}
\resizebox{\linewidth}{!}{\begin{tabular}{cccc cccc cccc cccc |l}
\toprule
 & \rotatebox[origin=c]{45}{Gaussian} & \rotatebox[origin=c]{45}{Shot} & \rotatebox[origin=c]{45}{Impluse} & \rotatebox[origin=c]{45}{Defocus} & \rotatebox[origin=c]{45}{Glass} & \rotatebox[origin=c]{45}{Motion} & \rotatebox[origin=c]{45}{Zoom} & \rotatebox[origin=c]{45}{Frost} & \rotatebox[origin=c]{45}{Snow} & \rotatebox[origin=c]{45}{Fog} & \rotatebox[origin=c]{45}{Brightness} & \rotatebox[origin=c]{45}{Contrast} & \rotatebox[origin=c]{45}{Elastic} & \rotatebox[origin=c]{45}{Pixelate} & \rotatebox[origin=c]{45}{JEPG} & \rotatebox[origin=c]{45}{Avg.}\\
\midrule
Source & 48.0 & 47.1 & 48.3 & 32.4 & 14.4 & 39.3 & 39.2 & 55.7 & 50.1 & 51.5 & 75.1 & 52.0 & 29.5 & 48.1 & 60.3 & 46.1\\
DiffPure &  42.6 & 42.0 & 40.9 & 16.8 & 22.2 & 19.1 & 23.0 & 39.0 & 29.1 & 9.1 & 62.5 & 3.0 & 34.3 & 46.8 & 60.7 & 32.7 \\  
DDA & 59.9 & 55.7 & 55.9 & 30.7 & 30.5 & 38.1 & 39.2 & 53.2 & 48.0 & 41.9 & 71.0 & 48.1 & 46.9 & 59.5 & 62.0 & 49.4\\
\rowcolor{gray!20}SDA (Ours) & 60.3 & 57.6 &58.1 &35.4 & 35.3 & 42.9 & 42.4 & 55.5 & 49.9 & 44.6 & 72.6 & 45.4 & 50.2 & 63.0 & 64.5 & \textbf{51.9 \color{teal}{(+2.5)}}\\
\bottomrule 
\end{tabular}}\vspace{-1mm}
\caption{\textbf{Detailed comparisons of SDA and baselines across 15 adaptation domains of ImageNet-C.} SDA shows the best average accuracy. The results are tested with ConvNeXt-B. Comparisons using other models are deferred to the supplementary materials.}\label{tab:expinc-detail}\vspace{-2mm}
\end{table*}

\subsection{Main Results on ImageNet Classifiers}\label{sec:expin}

\textbf{Settings.} We choose DDA~\cite{dda} as our primary competitor since it is the best open-sourced method. SDA is also compared with DiffPure~\cite{diffpure}, MEMO~\cite{memo}, and the recent SOTA GDA~\cite{gda} using their reported results. Source model performance is reported as "Source". DiT~\cite{dit} and ADM~\cite{diffusion-beats-gans} are adopted to generate and align 50K synthetic data for 15-epoch finetuning. For different source models, the synthetic data only needs to be generated once.
Our results are tested on standard TTA benchmarks, ImageNet-C~\cite{inc} (severity level 5) and ImageNet-W~\cite{inw} using various models including ResNet~\cite{resnet}, Swin~\cite{swin} and ConvNeXt~\cite{convnext}. More implementation details are provided in the supplementary materials.

\vspace{1mm}
\noindent\textbf{Comparison results on ImageNet-C.} We begin by evaluating the performance of SDA on ImageNet-C. As reported in \cref{tab:expinc}, our proposed SDA consistently outperforms all baseline methods across different model architectures and sizes. We emphasize the performance improvement over DDA, as we adopted DDA for target data adaptation in SDA. Compared to DDA, our SDA improves accuracy by 2.5\%-2.9\%. This significant improvement indicates the misalignment between the source and synthetic domains, validating the effectiveness of our synthetic-domain alignment framework. Moreover, compared to the recent SOTA GDA, SDA also achieves an improvement of 2.2\% with ConvNeXt-T. Notably, SDA focuses on synthetic domain alignment, an orthogonal research direction to existing diffusion-driven methods on better adapting the target data. Therefore, the performance of SDA could potentially be further enhanced with the release of more advanced codebases like GDA. Compared to the model adaptation method, MEMO, three diffusion-driven methods (SDA, DDA, and GDA) all demonstrate superior performance, highlighting the effectiveness of diffusion models in assisting TTA tasks. DiffPure presents worse results since it is primarily designed for adversarial attacks. Without the structural guidance introduced in DDA and GDA, DiffPure may not effectively recover images with severe domain shifts.
In \cref{tab:expinc-detail}, we provide a detailed comparison of the results of SDA and baselines. SDA surpasses DiffPure in all 15 adaptation domains and outperforms DDA in 14 out of 15 domains, further affirming the superiority of SDA.

\begin{table}[t]
\centering
\small
\resizebox{\linewidth}{!}{\begin{tabular}{ccc|c>{\columncolor{gray!20}}c}
\toprule
 Model& Source & DiffPure  & DDA   & SDA (Ours) \\
\midrule
ResNet-50    &37.7 & 29.1  &52.8 & \textbf{54.7 \color{teal}{(+1.9)}}    \\
Swin-T        &66.5 & 52.7 &65.9 & \textbf{67.3 \color{teal}{(+1.4)}}\\
ConvNeXt-T  &  67.6 & 55.8  &67.9 &  \textbf{69.4 \color{teal}{(+1.5)}}    \\
Swin-B         & 69.1        &  55.5    &   68.3 &   \textbf{70.6 \color{teal}{(+2.3)}}      \\
ConvNeXt-B     & 70.1       &  57.7    & 70.3 & \textbf{72.3 \color{teal}{(+2.0)}} \\  
\bottomrule 
\end{tabular}}\vspace{-0mm}
\caption{\textbf{Quantitative results on ImageNet-W}. SDA shows consistent performance improvements compared to baselines.}\label{tab:expinu}\vspace{-3mm}
\end{table}

\begin{table*}[h]
\centering
\small
\setlength{\tabcolsep}{3pt}
\resizebox{\linewidth}{!}{\begin{tabular}{cccc cccc cccc cccc |l}
\toprule
 & \rotatebox[origin=c]{45}{Gaussian} & \rotatebox[origin=c]{45}{Shot} & \rotatebox[origin=c]{45}{Impluse} & \rotatebox[origin=c]{45}{Defocus} & \rotatebox[origin=c]{45}{Glass} & \rotatebox[origin=c]{45}{Motion} & \rotatebox[origin=c]{45}{Zoom} & \rotatebox[origin=c]{45}{Frost} & \rotatebox[origin=c]{45}{Snow} & \rotatebox[origin=c]{45}{Fog} & \rotatebox[origin=c]{45}{Brightness} & \rotatebox[origin=c]{45}{Contrast} & \rotatebox[origin=c]{45}{Elastic} & \rotatebox[origin=c]{45}{Pixelate} & \rotatebox[origin=c]{45}{JEPG} & \rotatebox[origin=c]{45}{Avg.}\\
\midrule
Source & 23.5 & 28.7 & 20.8 & 41.2 & 30.7 & 48.2 & 48.1 & 51.8 & 63.6 & 63.4 & 79.5 & 28.0 & 56.5 & 35.9 & 60.7 & 45.4 \\
DDA & 79.4 & 80.3 & 76.8 & 41.3 & 56.6 & 49.0 & 48.9 & 77.4 & 73.5 & 62.6 & 80.5 & 61.5 & 61.2 & 57.6 & 73.2 & 65.3\\
\rowcolor{gray!20}SDA (Ours)& 82.6 & 83.5 & 79.5 & 59.6 & 61.4 & 65.3 & 64.6 & 76.5 & 78.4 & 71.5 & 88.9 & 53.4 & 75.4 & 64.4 & 80.7 & \textbf{72.4 \color{teal}{(+7.1)}} \\

\bottomrule 
\end{tabular}}\vspace{-2mm}
\caption{\textbf{Comparisons of SDA and baselines with ResNet-18 on CIFAR-10-C.} SDA shows the best average accuracy.}\label{tab:cifar}\vspace{-2mm}
\end{table*}

\begin{table*}[t]
\centering
\small
\setlength{\tabcolsep}{3pt}
\resizebox{\linewidth}{!}{\begin{tabular}{cccc cccc cccc cccc |l}
\toprule
 & \rotatebox[origin=c]{45}{Gaussian} & \rotatebox[origin=c]{45}{Shot} & \rotatebox[origin=c]{45}{Impluse} & \rotatebox[origin=c]{45}{Defocus} & \rotatebox[origin=c]{45}{Glass} & \rotatebox[origin=c]{45}{Motion} & \rotatebox[origin=c]{45}{Zoom} & \rotatebox[origin=c]{45}{Frost} & \rotatebox[origin=c]{45}{Snow} & \rotatebox[origin=c]{45}{Fog} & \rotatebox[origin=c]{45}{Brightness} & \rotatebox[origin=c]{45}{Contrast} & \rotatebox[origin=c]{45}{Elastic} & \rotatebox[origin=c]{45}{Pixelate} & \rotatebox[origin=c]{45}{JEPG} & \rotatebox[origin=c]{45}{Avg.}\\
\midrule
Source & 9.8 & 13.2 & 10.0 & 22.7 & 5.7 & 20.4 & 22.1 & 36.7 & 28.4 & 53.9 & 66.6 & 9.7 & 24.3 & 16.0 & 43.8 & 25.6\\
DDA & 16.6 & 21.1 & 16.0 & 45.0 & 40.9 & 49.7 & 38.7 & 36.7 & 38.6 & 35.0 & 55.2 & 13.3 & 56.6 & 56.3 & 58.8 & 38.6\\
\rowcolor{gray!20}SDA (Ours) & 17.0& 21.1& 16.2 & 45.5&43.0& 51.1& 39.9& 38.7 & 40.2 &36.3 &57.9& 13.1 & 59.1 & 58.1 & 59.7 &\textbf{39.8 \color{teal}{(+1.2)}}\\
\bottomrule 
\end{tabular}}\vspace{-2mm}
\caption{\textbf{Comparisons of SDA and baselines with DeepLabv3 on PASCAL VOC-C.} SDA shows the best average mIOU.}\label{tab:seg}\vspace{-3mm}
\end{table*}

\begin{table*}[t]
\centering
\small
\setlength{\tabcolsep}{3pt}
\resizebox{\linewidth}{!}{\begin{tabular}{cccc cccc cccc cccc |l}
\toprule
 & \rotatebox[origin=c]{45}{Gaussian} & \rotatebox[origin=c]{45}{Shot} & \rotatebox[origin=c]{45}{Impluse} & \rotatebox[origin=c]{45}{Defocus} & \rotatebox[origin=c]{45}{Glass} & \rotatebox[origin=c]{45}{Motion} & \rotatebox[origin=c]{45}{Zoom} & \rotatebox[origin=c]{45}{Frost} & \rotatebox[origin=c]{45}{Snow} & \rotatebox[origin=c]{45}{Fog} & \rotatebox[origin=c]{45}{Brightness} & \rotatebox[origin=c]{45}{Contrast} & \rotatebox[origin=c]{45}{Elastic} & \rotatebox[origin=c]{45}{Pixelate} & \rotatebox[origin=c]{45}{JEPG} & \rotatebox[origin=c]{45}{Avg.}\\
\midrule
Source-Zero & 75.2 & 76.6 & 79.6 & 81.4 & 74.5 & 83.5 & 81.1 & 87.0 & 88.9 & 89.5 & 95.6 & 84.1 & 76.2 & 90.1 & 88.7 & 83.5 \\
Source & 86.8 & 86.0 & 88.6 & 90.1 & 85.5 & 93.5 & 89.7 & 93.6 & 95.0 & 94.9 & 98.1 & 89.5 & 85.6 & 96.2 & 94.8 & 91.2\\
DDA & 92.4 & 91.5 & 92.5 & 88.6 & 90.1 & 92.2 & 88.5 & 93.0 & 93.1 & 89.6 & 97.8 & 75.8 & 89.6 & 96.4 & 97.0 & 91.2 \\
\rowcolor{gray!20}SDA (Ours)& 94.9 & 94.1 & 94.6 & 91.0 & 92.9 & 94.0 & 91.5 & 93.6 & 95.2 & 92.2 & 98.5 & 83.6 & 92.8 & 97.3 & 97.6 & \textbf{93.6 \color{teal}{(+2.4)}}  \\

\bottomrule 
\end{tabular}}\vspace{-2mm}
\caption{\textbf{Comparisons of SDA and baselines with LLaVA on ImageNet-C.} SDA shows the best average accuracy.}\label{tab:llava}\vspace{-2mm}
\end{table*}

\vspace{1mm}
\noindent\textbf{Comparison results on ImageNet-W.}
We extend our evaluation to ImageNet-W to assess SDA's performance under watermark-based domain shifts. As shown in~\cref{tab:expinu}, SDA consistently surpasses all baselines across different models. Compared to our primary baseline, DDA, SDA achieves accuracy gains ranging from 1.4\% to 2.3\%. Furthermore, the results in~\cref{tab:expinu} reveal potential performance drops when DDA is applied to ImageNet-W with Swin-T and Swin-B models, suggesting that synthetic data may be less recognizable by the original source models. Although advancements in diffusion techniques could potentially improve outcomes, the consistent gains achieved by SDA indicate that aligning the source model with the synthetic domain offers a convenient and effective solution to enhance performance.

\subsection{Scalability to Other Datasets, Tasks and Models}\label{sec:expscale}

In addition to standard benchmarks, an effective TTA method should demonstrate its superiority across various dataset scales, task formats, and model architectures. In this section, we assess SDA from these aspects to validate its scalability. Implementation details of each experiment can be found in the supplementary materials.

\vspace{1mm}
\noindent\textbf{Scaling to small datasets.} 
We first evaluate the effectiveness of SDA on scenarios where both source and target domains are small-scale. Specifically, we test SDA on CIFAR-10-C~\cite{inc} with ResNet-18~\cite{resnet} as the source classifier. EDM~\cite{edm} is used to generate synthetic data, and I-DDPM~\cite{improved-diffusion} is applied for data alignment. As shown in~\cref{tab:cifar}, SDA consistently outperforms both the source model and DDA, achieving an average accuracy improvement of 7.1\% over DDA.

\vspace{1mm}
\noindent\textbf{Scaling to Semantic Segmentation Tasks.} We extend our evaluation to dense prediction tasks by using PASCAL-VOC-C~\cite{voc} as a standard semantic segmentation benchmark with DeepLabv3~\cite{deeplab} as the source segmenter. Dataset Diffusion~\cite{datasetdiffusion} generates synthetic segmentation data, and FLUX Schnell~\cite{flux} is used for data alignment. As shown in~\cref{tab:seg}, SDA achieves the best performance, with an average mIOU improvement of 1.2\% over DDA.

\vspace{1mm}
\noindent\textbf{Scaling to multimodal large language models (MLLMs).} As an emerging research direction, MLLMs~\cite{llava} present advanced visual question answering~\cite{vqa} capability. We design a language-based classification task format to test how SDA can help MLLMs on TTA tasks on ImageNet-C: Given an image, ask an MLLM (LLaVA 1.5-7b~\cite{llava} in our experiments) to choose the correct image class from four provided options. In~\cref{tab:llava}, the ``Source-Zero'' setting tests the zero-shot results of pretrained LLaVA. The ``Source'' and ``DDA'' settings evaluate a source-data fine-tuned LLaVA while the ``SDA'' setting tests a synthetic-data fine-tuned LLaVA. The fine-tuning task format is the same as that at test time. Pretrained LLaVA already exhibits strong performance. While DDA improves results in some domains, it does not yield an overall gain compared to source-data fine-tuned LLaVA. In contrast, SDA aligns LLaVA with the synthetic domain, achieving the best accuracy with an improvement of 2.4\%.

\begin{table*}[h]
\centering
\small
\setlength{\tabcolsep}{3pt}
\resizebox{\linewidth}{!}{\begin{tabular}{c|ccc cccc ccc |c >{\columncolor{gray!20}}c}
\toprule
\rotatebox[origin=c]{45}{Source} & \rotatebox[origin=c]{45}{TENT~\cite{tent}} & \rotatebox[origin=c]{45}{ROID~\cite{roid}} & \rotatebox[origin=c]{45}{NOTE~\cite{note}} & \rotatebox[origin=c]{45}{CoTTA~\cite{cotta}} & \rotatebox[origin=c]{45}{TRIBE~\cite{tribe}} & \rotatebox[origin=c]{45}{BN~\cite{bn3}} & \rotatebox[origin=c]{45}{UnMIX~\cite{umix}} &  \rotatebox[origin=c]{45}{RoTTA~\cite{rotta}} & \rotatebox[origin=c]{45}{LAME~\cite{lame}} & \rotatebox[origin=c]{45}{UniTTA~\cite{unitta}} &   \rotatebox[origin=c]{45}{DDA~\cite{dda}} & \rotatebox[origin=c]{45}{SDA (Ours)}\\
\midrule
18.3 & 4.2 & 6.2 & 6.3 & 8.8 & 9.3 & 9.8 & 17.4 & 23.1 & 24.9 & 30.3  & 29.3 & \textbf{31.6 \color{teal}{(+2.3)}} \\

\bottomrule 
\end{tabular}}\vspace{-2mm}
\caption{\textbf{Data stream sensitivity comparison.} 
We additionally compare SDA with 10 traditional TTA methods on the UniTTA~\cite{unitta} benchmark which contains 12 class/domain balance/imbalance settings. Results are reported as average accuracy across settings.}\label{tab:unitta}\vspace{-4mm}
\end{table*}

\subsection{Analysis}\label{sec:expan}

\begin{figure}[h]
  \centering
  \includegraphics[width=\linewidth]{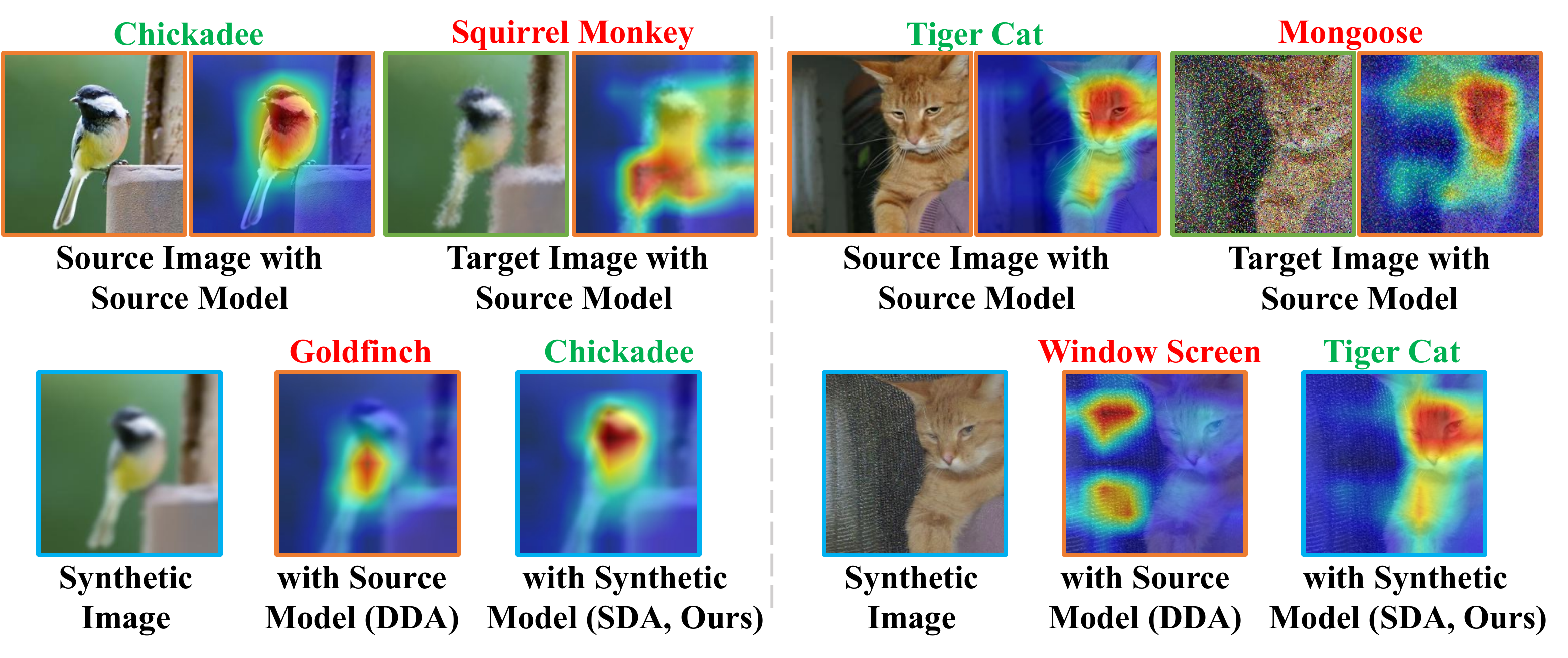}\vspace{-2mm}
 \caption{\textbf{Grad-CAM visualization comparison.} The first row shows activation maps for \textit{source} and \textit{target} images tested with the \textit{source} model. The second row displays activation maps for diffusion \textit{synthetic} images tested with the \textit{source} model (DDA) and our \textit{synthetic}-domain model (SDA). SDA aligns closely with the \textit{source} model’s response to \textit{source} images.}\vspace{-4mm}
  \label{fig:vis}
\end{figure}

\vspace{1mm}
\noindent\textbf{Visualization.} To demonstrate how synthetic data fine-tuning in SDA enhances the performance of diffusion-driven TTA methods, we employ Gradient-weighted Class Activation Mapping 
 (Grad-CAM)~\cite{gradcam} to visualize the image regions that most influence classification scores across different images and models. As shown in~\cref{fig:vis}, testing target images with the source model reveals distinct differences in activation maps and the occurrence of incorrect predictions compared to those from source images, underscoring the performance degradation due to domain shifts. Despite using adapted synthetic images, DDA still risks focusing on inappropriate regions and producing incorrect predictions. This highlights the domain misalignment of the synthetic data and source model. In contrast, SDA aligns both the data and model within the same synthetic domain, thereby producing activation maps and predictions that closely resemble those produced by the source model on source images.

\vspace{1mm}
\noindent\textbf{Data stream sensitivity.}
In~\cref{tab:unitta}, we test data stream sensitivity using the UniTTA~\cite{unitta} benchmark which contains 12 class/domain balance/imbalance settings closely aligned with diverse real-world TTA scenarios. In addition to diffusion-driven TTA methods, we include 10 additional popular traditional TTA methods~\cite{tent,roid,note,cotta,tribe,bn3,umix,rotta,lame,unitta} for comparison. We report the average accuracy of 12 settings using ResNet-50. The results indicate that diffusion-driven TTA methods are preferred in these challenging settings since they are insensitive to different variants of data streams. SDA maintains this insensitivity and showcases the best performance. Detailed comparisons for each setting are provided in the supplementary materials.

\subsection{Ablation Studies}\label{expab}

\textbf{Components.} We examine the impact of two key components in our SDA framework's Mix of Diffusion technique, as outlined in \cref{table:abc}: (1) synthetic data generation using the conditional diffusion model and (2) synthetic data alignment using the unconditional diffusion model. Fine-tuning source models with synthetic data generated solely by the conditional diffusion model (+ Conditional Data Generation) yields only marginal improvements. As discussed in \cref{sec:mod}, this limited gain arises from a domain misalignment between the conditional and unconditional diffusion models. Specifically, using only conditional synthetic data results in models aligned with the conditional diffusion domain, whereas the test data belongs to the unconditional diffusion domain.
Therefore, further aligning the synthetic data through the unconditional diffusion model (+ Unconditional Data Alignment) leads to significant performance gains, surpassing the baseline DDA. This demonstrates that bridging the misalignment of different diffusion domains is essential for the success of our SDA framework.

\begin{table}[t]
\centering
\small
\setlength{\tabcolsep}{3pt}
\resizebox{\linewidth}{!}{\begin{tabular}{lccc}
    \toprule
      Component& ResNet-50   & Swin-T & ConvNeXt-T \\
    \midrule
    DDA~\cite{dda} & 29.7 & 40.0 & 44.2 \\
    + Conditional Data Generation &30.4 &40.1 & 44.6 \\
    \rowcolor{gray!20}+ Unconditional Data Alignment &\textbf{32.5}  &\textbf{42.5}  & \textbf{47.0} \\
    \bottomrule
\end{tabular}}\vspace{-2mm}
\caption{\textbf{Impact of different components in SDA.} Results are evaluated on ImageNet-C.}\label{table:abc}\vspace{-4mm}
\end{table}

\vspace{1mm}
\noindent\textbf{Number of fine-tuning images.} We examine the impact of different numbers of images ($N$) used during synthetic-domain model fine-tuning in \cref{table:abn}. Interestingly, even with only one image per class ($N$ = 1K), SDA still significantly outperforms DDA. This finding suggests a key attribute of the fine-tuning process: source models are primarily learning to adapt to the synthetic domain itself, rather than acquiring class-specific knowledge. Increasing the number of images helps the fine-tuning process capture the synthetic domain more accurately, thereby enhancing performance. Based on a balance between performance improvement and image generation resources, we select $N$ = 50K as our default experimental setting.

\begin{table}[t]
\centering
\small
\setlength{\tabcolsep}{3pt}
\resizebox{1\linewidth}{!}{\begin{tabular}{ccccc}
    \toprule
    {Method}& {Number ($N$)}&  ResNet-50   & Swin-T & ConvNeXt-T \\
    \midrule
DDA~\cite{dda}&0 & 29.7 & 40.0 & 44.2 \\
\midrule
\multirow{4}{*}{SDA}&1K & 31.9 &42.3 &45.5 \\
&10K & 31.9 &\textbf{42.5}& 46.6 \\
&50K (default) & 32.5 & \textbf{42.5} & \textbf{47.0} \\
&100K & \textbf{32.6} & 42.2& 46.8 \\
    \bottomrule
\end{tabular}}\vspace{-2mm}
\caption{\textbf{Impact of different numbers of fine-tuning images.} Results are evaluated on ImageNet-C.}\label{table:abn}\vspace{-4mm}
\end{table}



\section{Conclusion}
\label{sec:con}

In this paper, we proposed \textbf{S}ynthetic-\textbf{D}omain \textbf{A}lignment (\textbf{SDA}), a novel test-time adaptation (TTA) framework that simultaneously aligns the domains of the source model and target data with the synthetic domain of a diffusion model. For the source model, SDA introduces a Mix of Diffusion (MoD) technique, which generates synthetic data to adapt the source model to a synthetic-domain model. 
MoD involves a conditional diffusion model for data generation and an unconditional diffusion model for data alignment. 
For the target data, SDA utilizes the aforementioned unconditional diffusion model to project the target data to synthetic data. As the domains of the model and data are aligned, SDA converts the cross-domain TTA task into an easier in-domain prediction task. Compared to existing diffusion-driven TTA methods, SDA significantly mitigates the source-synthetic domain misalignment issue. Compared to traditional TTA methods, SDA maintains insensitivity to different data streams. Extensive experiments across classifiers, segmenters, and MLLMs indicate that SDA achieves enhanced domain alignment and superior performance.



{
    \small
    \bibliographystyle{ieeenat_fullname}
    \bibliography{main}
}

\appendix
\clearpage
\setcounter{page}{1}
\maketitlesupplementary

\section{Implementation Details}

\subsection{Baselines.}
We choose DDA~\cite{dda} as our primary competitor since it is the best-performing publicly available diffusion-driven TTA method. Same as DDA, we include DiffPure~\cite{diffpure} and MEMO~\cite{memo} as baselines. We also compare SDA against the recent SOTA GDA~\cite{gda} using their paper results. For data stream sensitivity comparison, we compare SDA with 10 additional traditional TTA methods, including TENT~\cite{tent}, ROID~\cite{roid}, NOTE~\cite{note}, CoTTA~\cite{cotta}, TRIBE~\cite{tribe}, BN~\cite{bn3}, UniMIX~\cite{umix}, RoTTA~\cite{rotta}, LAME~\cite{lame} and UniTTA~\cite{unitta}. The results are evaluated across various TTA benchmarks, including ImageNet-C~\cite{inc}, ImageNet-W~\cite{inw}, CIFAR-10-C~\cite{inc} and PASCAL VOC-C~\cite{voc}.

\subsection{Settings.} 
\begin{figure}[t]
  \centering
  \includegraphics[width=0.8\linewidth]{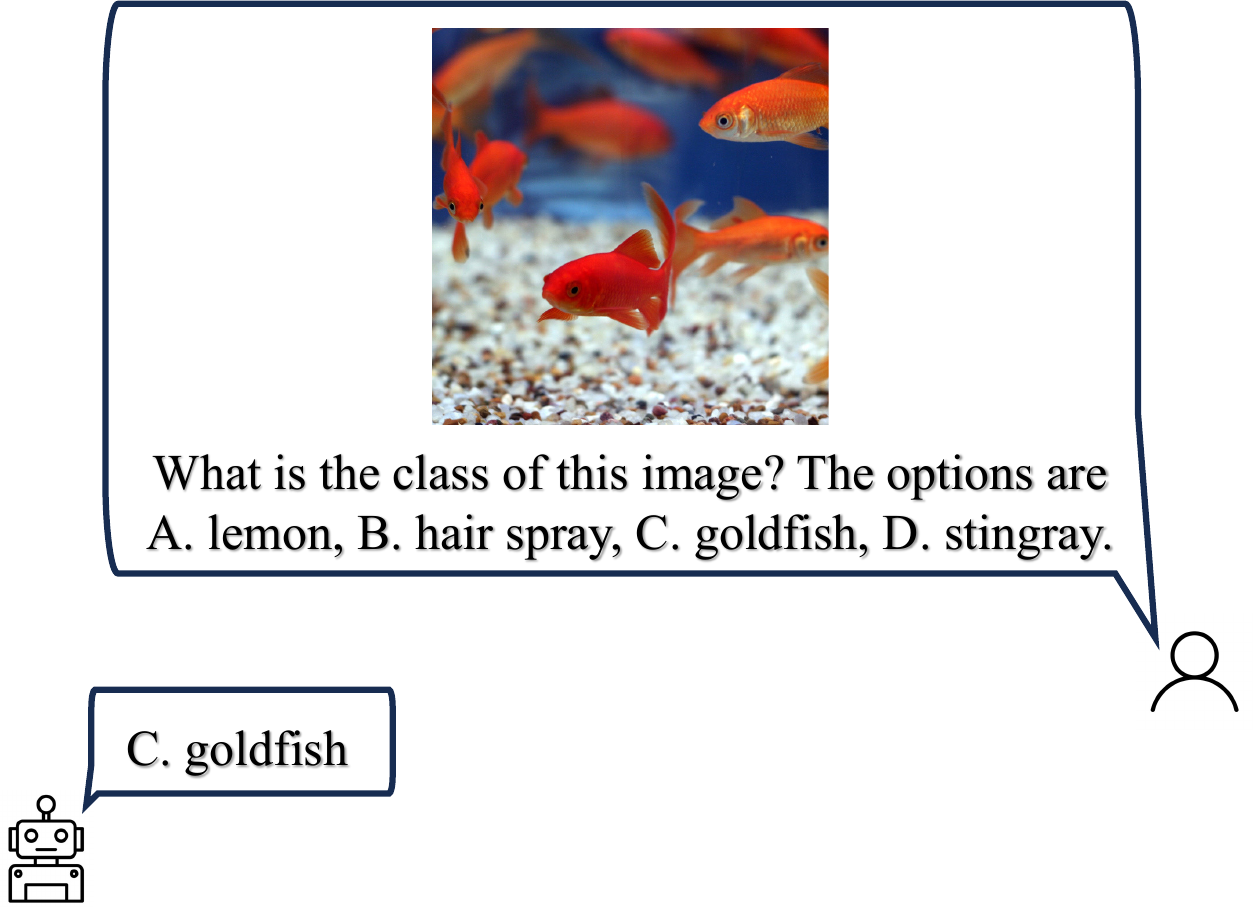}
  \vspace{-2mm}
  \caption{Task format for fine-tuning and evaluating MLLMs. Given an image, we ask an MLLM to choose the correct image class from four provided options.}
  \label{fig:task}
\end{figure}

All experiments are conducted with 8 A100 GPUs. For ImageNet variants, we explore ResNet~\cite{resnet}, ConvNeXt~\cite{convnext}, and Swin~\cite{swin} as source models. DiT~\cite{dit} and ADM~\cite{diffusion-beats-gans} are adopted as conditional and unconditional diffusion models, respectively. For CIFAR-10-C~\cite{inc}, we use ResNet as the source model. EDM~\cite{edm} and I-DDPM~\cite{improved-diffusion} are adopted as conditional and unconditional diffusion models, respectively. For PASCAL VOC-C~\cite{voc}, we use DeepLabv3~\cite{deeplab} as the source segmenter. Dataset Diffusion~\cite{datasetdiffusion} and FLUX schnell~\cite{flux} are adopted as conditional and unconditional diffusion models, respectively. For classification tasks via MLLMs, we use LLaVA 1.5-7b~\cite{llava} as the source model.  For each task, we generate 50K images with balanced class labels. For different source models and target domains, the synthetic data only needs to be generated once.
The detailed fine-tuning settings of classifiers and segmenters are summarized in~\cref{table:config}. For MLLM (LLaVA) fine-tuning, we follow the default configurations in~\cite{llava}. \cref{fig:task} shows the task format for fine-tuning and evaluating MLLMs.

\section{Selection of Timestep for TTA} 
\begin{figure}[t]
  \centering
  \includegraphics[width=\linewidth]{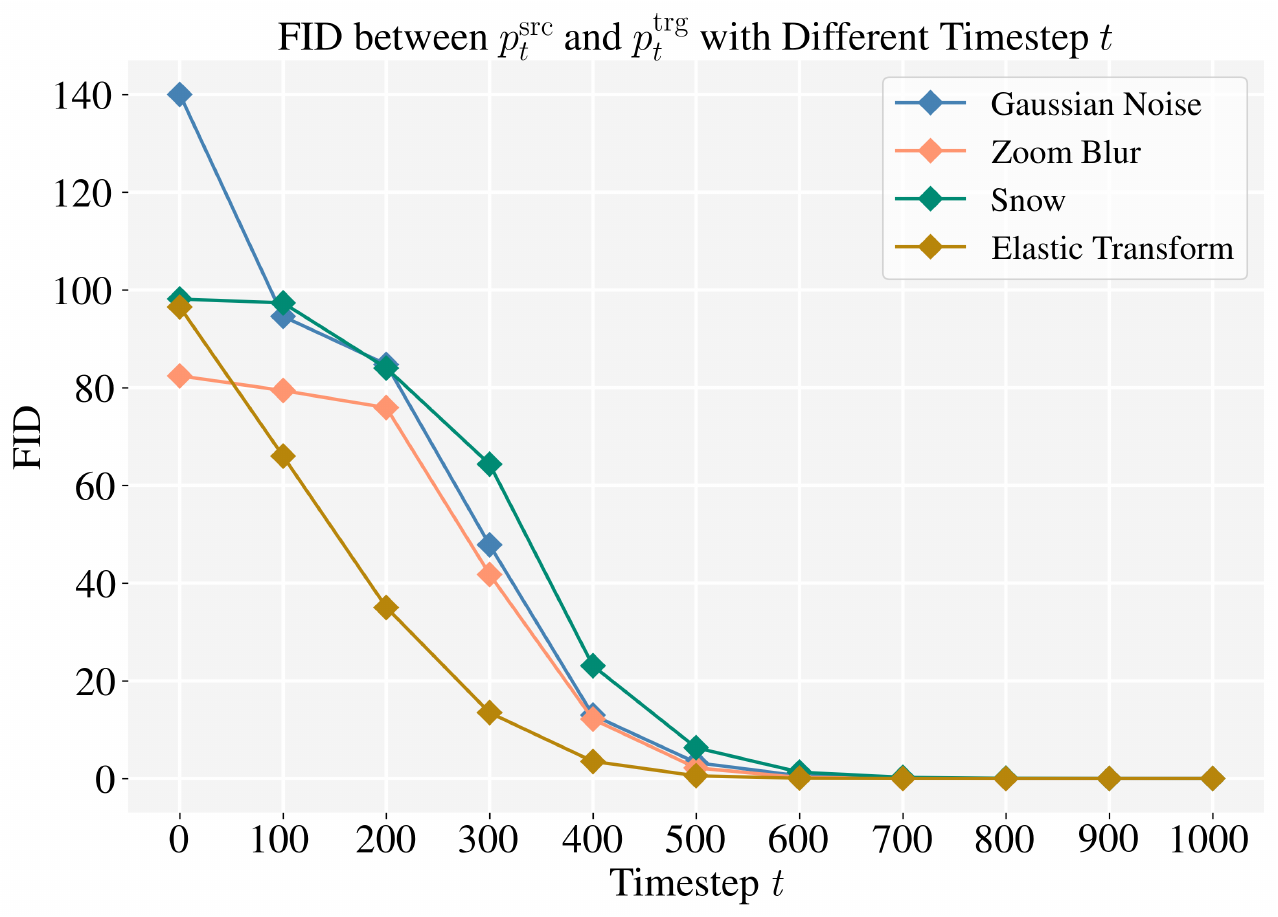}
  \vspace{-6mm}
  \caption{Fréchet Inception Distance (FID)~\cite{fid} between $p_{t}^{\rm src}$ and $p_{t}^{\rm trg}$ with different timestep $t$. We conduct experiments on four typical adaptation types from ImageNet-C.}\vspace{-2mm}
  \label{fig:fid}
\end{figure}

As aforementioned in Eq.~\ref{eq:t*}, the success of diffusion-driven data adaptation relies on the selection of a suitable minimum $t^*$ that satisfies $p_{t^*}^{\rm src} \approx p_{t^*}^{\rm trg}$. In~\cref{fig:fid}, we leverage FID~\cite{fid} to measure the domain divergence of $p_{t}^{\rm src}$ and $p_{t}^{\rm trg}$ with different timestep $t$. The results indicate that for a 1000-step diffusion scheduler and adaptation tasks from the standard benchmark ImageNet-C~\cite{inc}, diffusion-driven data adaptation typically requires a $t^*$ larger than 500. We empirically demonstrate that applying such $t^*$ to diffusion-driven TTA methods leads to significant misalignment between the source and synthetic domains, as shown in~\cref{tab:pre}. In our experiments, we set the same $t^*=500$ as our baseline DDA~\cite{dda}. Here $t^*=500$ refers to using half sampling steps as the whole diffusion scheduler, \eg, for a 100-step scheduler, the actual sampling step for adaptation is 50.

\section{Additional Results} We provide detailed comparisons of SDA and baselines across 15 adaptation domains of ImageNet-C in~\cref{tab:sb,tab:ct,tab:st,tab:r50} and across 12 class/domain balance/imbalance settings from the UniTTA benchmark~\cite{unitta} in~\cref{tab:unitta-supp}.

\begin{table*}[h]
\centering
\small
\setlength{\tabcolsep}{3pt}
\resizebox{\linewidth}{!}{\begin{tabular}{l|cccc}
    \toprule
    Dataset & \multicolumn{2}{c}{ImageNet} & CIFAR-10 & PASCAL VOC \\
    \midrule
     Model & ResNet-50   & Swin-T/B \& ConvNeXt-T/B & ResNet-18 & DeepLabv3 \\
    \midrule
    optimizer & SGD & AdamW & SGD & SGD\\
    base learning rate & 5e-4 & 2e-5 & 5e-2 & 1e-4 \\
    weight decay & 1e-4 & 1e-8 & 1e-4 & 5e-6\\
    optimizer momentum & 0.9 & $\beta_1,\beta_2=0.9, 0.999$ 
 & 0.9 & 0.9\\
    batch size & 512 & 1024 & 128 & 32\\
    training epochs & 15 & 15 & 15 & 2500 (iterations)\\
    learning rate schedule & step decay at epoch 10 & cosine decay & step decay at epoch 10 & polynomialLR\\
    warmup epochs & None & 5 & None & None\\
    warmup schedule & N/A & linear & N/A & N/A\\
    conditional diffusion model & DiT-XL/2 & DiT-XL/2 & EDM-VP & Dataset Diffusion\\
    conditional sampling steps & 250 & 250 & 512 & 100 \\
    classifier-free guidance & 1.0 & 1.0 & 1.0 & 7.5  \\
    unconditional diffusion model & ADM & ADM & I-DDPM & FLUX schnell\\
    unconditional sampling steps & 50 & 50 & 50 & 25 \\
    \bottomrule
\end{tabular}}
\vspace{-2mm}
\caption{Synthetic-domain model adaptation settings.}\label{table:config}
\end{table*}

\begin{table*}[t]
\centering
\small
\vspace{-1mm}
\setlength{\tabcolsep}{3pt}
\resizebox{\linewidth}{!}{\begin{tabular}{cccc cccc cccc cccc |l}
\toprule
 & \rotatebox[origin=c]{45}{Gaussian} & \rotatebox[origin=c]{45}{Shot} & \rotatebox[origin=c]{45}{Impluse} & \rotatebox[origin=c]{45}{Defocus} & \rotatebox[origin=c]{45}{Glass} & \rotatebox[origin=c]{45}{Motion} & \rotatebox[origin=c]{45}{Zoom} & \rotatebox[origin=c]{45}{Frost} & \rotatebox[origin=c]{45}{Snow} & \rotatebox[origin=c]{45}{Fog} & \rotatebox[origin=c]{45}{Brightness} & \rotatebox[origin=c]{45}{Contrast} & \rotatebox[origin=c]{45}{Elastic} & \rotatebox[origin=c]{45}{Pixelate} & \rotatebox[origin=c]{45}{JEPG} & \rotatebox[origin=c]{45}{Avg.}\\
\midrule
Source & 39.1 & 37.7 & 38.8 & 29.0 & 11.1 & 33.4 & 34.7 & 51.1 & 43.4 & 59.8 & 71.3 & 41.2 & 27.1 & 35.9 & 54.0 & 40.5\\
DDA & 53.8 & 49.2 & 50.3 & 28.5 & 26.2 & 33.4 & 34.9 & 49.4 & 42.8 & 40.9 & 67.9 & 38.0 & 43.1 & 52.7 & 57.1 & 44.5\\
\rowcolor{gray!20}SDA (Ours) &55.3 & 53.5 & 53.7 & 32.5 & 31.1 & 37.7 & 38.3 & 51.1 & 43.8 & 42.4 & 69.7 & 34.4 & 47.8 & 58.3 & 60.8
 & \textbf{47.4 \color{teal}{(+2.9)}}\\
\bottomrule 
\end{tabular}}\label{tab:expinc-detail-sb}
\vspace{-2mm}
\caption{Comparisons of SDA and baselines across 15 adaptation domains of ImageNet-C. Results are conducted with Swin-B.}\label{tab:sb}
\end{table*}

\begin{table*}[t]
\centering
\small
\vspace{-1mm}
\setlength{\tabcolsep}{3pt}
\resizebox{\linewidth}{!}{\begin{tabular}{cccc cccc cccc cccc |l}
\toprule
 & \rotatebox[origin=c]{45}{Gaussian} & \rotatebox[origin=c]{45}{Shot} & \rotatebox[origin=c]{45}{Impluse} & \rotatebox[origin=c]{45}{Defocus} & \rotatebox[origin=c]{45}{Glass} & \rotatebox[origin=c]{45}{Motion} & \rotatebox[origin=c]{45}{Zoom} & \rotatebox[origin=c]{45}{Frost} & \rotatebox[origin=c]{45}{Snow} & \rotatebox[origin=c]{45}{Fog} & \rotatebox[origin=c]{45}{Brightness} & \rotatebox[origin=c]{45}{Contrast} & \rotatebox[origin=c]{45}{Elastic} & \rotatebox[origin=c]{45}{Pixelate} & \rotatebox[origin=c]{45}{JEPG} & \rotatebox[origin=c]{45}{Avg.}\\
\midrule
Source & 40.1 & 39.1 & 38.7 & 25.6 & 11.4 & 33.0 & 31.2 & 49.3 & 43.8 & 41.9 & 70.3 & 45.0 & 22.5 & 41.0 & 57.2 & 39.3\\
DDA & 55.6 & 51.6 & 51.3 & 24.7 & 26.9 & 31.9 & 32.3 & 48.4 & 42.6 & 34.3 & 66.7 & 39.9 & 42.2 & 54.6 & 59.3 & 44.2\\
\rowcolor{gray!20}SDA (Ours) & 56.7 & 53.9 & 53.8 & 29.9 & 32.0 & 36.2 & 36.8 & 49.7 & 43.7 & 36.4 & 68.0 & 39.0 & 47.1 & 59.8 & 62.1 & \textbf{47.0 \color{teal}{(+2.8)}}\\
\bottomrule 
\end{tabular}}\label{tab:expinc-detail-ct}\vspace{-2mm}
\caption{Comparisons of SDA and baselines across 15 adaptation domains of ImageNet-C. Results are conducted with ConNeXt-T.}\label{tab:ct}
\end{table*}

\begin{table*}[t]
\centering
\small
\vspace{-1mm}
\setlength{\tabcolsep}{3pt}
\resizebox{\linewidth}{!}{\begin{tabular}{cccc cccc cccc cccc |l}
\toprule
 & \rotatebox[origin=c]{45}{Gaussian} & \rotatebox[origin=c]{45}{Shot} & \rotatebox[origin=c]{45}{Impluse} & \rotatebox[origin=c]{45}{Defocus} & \rotatebox[origin=c]{45}{Glass} & \rotatebox[origin=c]{45}{Motion} & \rotatebox[origin=c]{45}{Zoom} & \rotatebox[origin=c]{45}{Frost} & \rotatebox[origin=c]{45}{Snow} & \rotatebox[origin=c]{45}{Fog} & \rotatebox[origin=c]{45}{Brightness} & \rotatebox[origin=c]{45}{Contrast} & \rotatebox[origin=c]{45}{Elastic} & \rotatebox[origin=c]{45}{Pixelate} & \rotatebox[origin=c]{45}{JEPG} & \rotatebox[origin=c]{45}{Avg.}\\
\midrule
Source & 29.9 & 28.2 & 28.3 & 23.1 & 9.5 & 24.4 & 27.8 & 46.6 & 36.3 & 47.0 & 68.4 & 34.5 & 20.8 & 27.4 & 50.1 & 33.5\\
DDA & 51.4 & 46.6 & 46.3 & 21.0 & 22.1 & 23.9 & 27.9 & 45.5 & 36.2 & 40.5 & 64.3 & 30.6 & 40.4 & 48.5 & 54.2 & 40.0\\
\rowcolor{gray!20}SDA (Ours) & 52.4 & 50.2 & 50.1 & 24.4 & 26.5 & 29.1 & 32.4 & 46.2 & 37.3 & 38.8 & 65.3 & 26.1 & 46.1 & 55.3 & 57.2 &
\textbf{42.5 \color{teal}{(+2.5)}}\\
\bottomrule 
\end{tabular}}\label{tab:expinc-detail-st}\vspace{-2mm}
\caption{Comparisons of SDA and baselines across 15 adaptation domains of ImageNet-C. Results are conducted with Swin-T.}\label{tab:st}
\end{table*}

\begin{table*}[t]
\centering
\small
\vspace{-1mm}
\setlength{\tabcolsep}{3pt}
\resizebox{\linewidth}{!}{\begin{tabular}{cccc cccc cccc cccc |l}
\toprule
 & \rotatebox[origin=c]{45}{Gaussian} & \rotatebox[origin=c]{45}{Shot} & \rotatebox[origin=c]{45}{Impluse} & \rotatebox[origin=c]{45}{Defocus} & \rotatebox[origin=c]{45}{Glass} & \rotatebox[origin=c]{45}{Motion} & \rotatebox[origin=c]{45}{Zoom} & \rotatebox[origin=c]{45}{Frost} & \rotatebox[origin=c]{45}{Snow} & \rotatebox[origin=c]{45}{Fog} & \rotatebox[origin=c]{45}{Brightness} & \rotatebox[origin=c]{45}{Contrast} & \rotatebox[origin=c]{45}{Elastic} & \rotatebox[origin=c]{45}{Pixelate} & \rotatebox[origin=c]{45}{JEPG} & \rotatebox[origin=c]{45}{Avg.}\\
\midrule
Source & 6.1 & 7.5 & 6.7 & 14.3 & 7.6 & 11.8 & 21.5 & 21.4 & 16.2 & 19.1 & 55.1 & 3.6 & 14.5 & 33.3 & 42.1 & 18.7\\
DDA & 46.9 & 42.0 & 41.3 & 13.8 & 16.4 & 12.0 & 22.3 & 26.8 & 21.0 & 17.1 & 51.1 & 3.1 & 36.2 & 45.7 & 50.2 & 29.7\\
\rowcolor{gray!20}SDA (Ours) & 43.4 & 43.2 & 42.5 & 18.8 & 21.6 & 16.6 & 27.4 & 30.0 & 22.6 & 18.1 & 53.1 & 3.1 & 41.0 & 52.1 & 53.4 & 
\textbf{32.5 \color{teal}{(+2.8)}}\\
\bottomrule 
\end{tabular}}\label{tab:expinc-detail-re}\vspace{-2mm}
\caption{Comparisons of SDA and baselines across 15 adaptation domains of ImageNet-C. Results are conducted with ResNet-50.}\label{tab:r50}
\end{table*}

\begin{table*}[h]
\centering
\small
\setlength{\tabcolsep}{3pt}
\resizebox{\linewidth}{!}{\begin{tabular}{c cc ccccc ccccc |l}
\toprule
Class setting & \multicolumn{2}{c}{i.i.d. and balanced (i,1)} & \multicolumn{5}{c}{non-i.i.d. and balanced (n,1)} & \multicolumn{5}{c}{non-i.i.d. and imbalanced (n,u)} & \\ 
\cmidrule(l){2-3} \cmidrule(l){4-8} \cmidrule(l){9-13}
Domain setting & (1,1) & (i,1) & (1,1) & (i,1) & (i,u) & (n,1) & (n,u) & (1,1) & (i,1) & (i,u) & (n,1) & (n,u) &  \\ 
\cmidrule{2-3} \cmidrule(l){4-8} \cmidrule(l){9-13}
Corresponding setting & CoTTA & ROID & RoTTA & - & - & - & - & TRIBE & - & - & - & - & Avg. \\ \midrule
Source & 18.01 & 17.95 & 18.08 & 17.90 & 18.34 & 18.04 & 18.26 & 18.40 & 18.79 & 18.58 & 18.80 & 18.48 & 18.30 \\ 
\midrule
TENT~\cite{tent} & 29.42 & 8.12 & 1.28 & 0.69 & 0.47 & 0.88 & 0.68 & 2.50 & 0.78 & 0.87 & 2.97 & 1.14 & 4.15 \\ 
ROID~\cite{roid} & \textbf{39.33} & 20.82 & 1.49 & 0.29 & 0.16 & 0.48 & 0.39 & 8.24 & 0.23 & 0.43 & 1.85 & 0.63 & 6.20 \\
NOTE~\cite{note} & 8.38 & 11.82 & 6.33 & 4.73 & 3.18 & 5.00 & 4.19 & 7.51 & 4.07 & 4.59 & 11.07 & 4.95 & 6.32 \\ 
CoTTA~\cite{cotta} & \underline{33.13} & 19.33 & 4.87 & 3.20 & 2.67 & 3.78 & 3.67 & 10.30 & 4.80 & 5.50 & 7.89 & 6.29 & 8.78 \\
TRIBE~\cite{tribe} & 24.12 & 15.22 & 10.22 & 7.38 & 3.46 & 4.81 & 4.01 & 11.28 & 7.15 & 6.29 & 10.63 & 5.95 & 9.21 \\
BN~\cite{bn3} & 30.67 & 17.13 & 6.21 & 4.92 & 4.85 & 4.90 & 4.99 & 11.60 & 7.76 & 7.75 & 8.69 & 8.16 & 9.80 \\ 
UnMIX-TNS~\cite{umix} & 20.36 & 14.45 & 20.26 & 15.58 & 17.33 & 15.43 & 17.19 & 21.33 & 16.72 & 17.66 & 14.96 & 17.62 & 17.40 \\
RoTTA~\cite{rotta} & 32.23 & 20.09 & 27.28 & 19.46 & 20.35 & 19.70 & 20.37 & 31.26 & 21.74 & 22.06 & 20.22 & 21.64 & 23.12 \\ 
LAME~\cite{lame} & 17.45 & 17.74 & 25.52 & 27.79 & \underline{28.23} & 26.48 & 26.87 & 24.30 & 26.56 & 26.46 & 25.62 & 25.61 & 24.88 \\
UniTTA~\cite{unitta} & 21.93 & 22.00 & 29.75 & \textbf{33.17} & \textbf{33.58} & \underline{31.71} & \textbf{31.95} & 27.98 & \textbf{34.32} & \textbf{33.13} & \underline{31.52} & \textbf{32.42} & \underline{30.29} \\ \midrule
DDA~\cite{dda} & 29.89 & \underline{30.32} & \underline{29.88} & 29.94 & 26.33 & 29.58 & 26.28 &\underline{31.67} & 31.28 & 27.29 & 31.3 & 28.18 & 29.33 \\
\rowcolor{gray!20}SDA (Ours) & 32.42 & \textbf{32.72} & \textbf{32.34} & \underline{32.50} & 27.75  & \textbf{32.06} & \underline{27.88} & \textbf{34.36} & \underline{34.05} & \underline{29.06} & \textbf{34.02} & \underline{29.99} & \textbf{31.60 \color{teal}{(+2.27)}} \\ 
\bottomrule
\end{tabular}}
\caption{Data stream sensitivity comparison on ImageNet-C~\cite{inc} under 12 class/domain balance/imbalance settings in the UniTTA benchmark~\cite{unitta}. Detailed introduction of the settings can be found in~\cite{unitta}. Briefly, (\{i, n, 1\}, \{1, u\})
denotes correlation and imbalance settings, where \{i, n, 1\} represent i.i.d., non-i.i.d. and continual,
respectively, and \{1, u\} represent balance and imbalance, respectively. The best results are in \textbf{bold} and the second-best results are \underline{underlined}.}
\label{tab:unitta-supp}
\end{table*}

\end{document}